# Improving Sentiment Analysis in Arabic Using Word Representation

Abdulaziz M. Alayba[1], Vasile Palade[2], Matthew England[3] and Rahat Iqbal[4]

*School of Computing, Electronics and Mathematics*
*Faculty of Engineering, Environment and Computing*
*Coventry University*
*Coventry, UK*

[1] Alaybaa@uni.coventry.ac.uk
[2,3,4] {Vasile.Palade, Matthew.England, R.Iqbal}@coventry.ac.uk

*Abstract* — **The complexities of Arabic language in morphology, orthography and dialects makes sentiment analysis for Arabic more challenging. Also, text feature extraction from short messages like tweets, in order to gauge the sentiment, makes this task even more difficult. In recent years, deep neural networks were often employed and showed very good results in sentiment classification and natural language processing applications. Word embedding, or word distributing approach, is a current and powerful tool to capture together the closest words from a contextual text.**

**In this paper, we describe how we construct Word2Vec models from a large Arabic corpus obtained from ten newspapers in different Arab countries. By applying different machine learning algorithms and convolutional neural networks with different text feature selections, we report improved accuracy of sentiment classification (91%-95%) on our publicly available Arabic language health sentiment dataset** [1].

## I. Introduction

Sentiment Analysis is one of the Natural Language Processing (NLP) tasks that deals with unstructured text and classifies it as expressing either a positive or a negative sentiment. There are also sentiment analysis tools to classify text into three classes (positive, neutral, negative) or more (e.g., very positive, positive, neutral, negative, and very negative). There has been a constant rise in the use of many social networks, such as TripAdvisor, Yelp, Foursquare, Booking, and Twitter. In such networks, users can write their opinions about services, food, places to visit, hotels, etc. These are rich resources, with huge numbers of opinions, represented as unstructured text in many different languages. Hence this data has gained much interest and focus from NLP researchers and has been widely explored in many languages, and especially in English.

Many techniques and approaches were used to improve NLP in general and sentiment analysis in particular, including machine learning algorithms, stemming and lemmatising the text, focusing on some words by using Part of Speech (POS) taggers, using lexicon based approaches, as well as by combining with word distributing techniques.

There is a growing body of research in NLP for the Arabic language in recent years, for example [2], [3] and [4]. However, there is still a need to tackle the complexity of NLP tasks in Arabic. This complexity comes from many aspects, such as morphology, orthography, dialects, short vowels and word order. For example, the Arabic letter Hamzah or Hamza (ء) can be written in four different forms ( أ ، ؤ ، ئ ، ء ), so people can easily make mistakes.

This paper will first overview some related recent works on sentiment analysis in Section II. Then, in Section III, we describe the details of a process of distributing Arabic words using the Word2Vec approach from an available Arabic corpus [5]. Section IV contains a brief description of the dataset, and the results of some machine learning algorithms along with convolutional deep neural networks that we used. It also presents different text feature selections and the way the features are used within the machine learning classifiers. Finally, Section V presents our conclusions and plans for future work.

## II. Related Work

Sentiment analysis gained exposure in [6], where three machine learning algorithms were used: Naïve Bayes, Maximum Entropy, and Support Vector Machines. Since [6] the amount of research on sentiment analysis has significantly increased. For example: [7] utilized semantic values to phrases and words as features; and [8] combined a lexicon with a Twitter followers' graph to help in the sentiment classification.

In recent years, attention to the Arabic language has also increased: [2] is a book for researchers dealing with Arabic NLP; [3] presented a rule-based approach for Arabic language using an adaption from other languages, such as English; [4] introduced a tool for preprocessing Arabic text, which include root stemmer, part-of-speech tagger (POS-tagger), etc.; and [9] reported some challenges in dealing with the Arabic language in NLP and described some solutions for these. Moreover, sentiment analysis in Arabic language has received individual attention: [10] used some machine learning methods for sentiment classification; [11] presented an annotated Arabic dataset and applied morphology-based and lexical features for Arabic sentiment analysis; [12] improved the performance of sentiment analysis for Arabic, using different techniques like stemming, POS and expanding lexicon; [13] used deep neural networks with three different architectures for Arabic sentiment analysis; and [14] considered building an Arabic lexicon, manually and automatically.

FIGURE I. WORD2VEC MODEL, CONTINUOUS BAG OF WORDS CBOW AND SKIP-GRAM SG

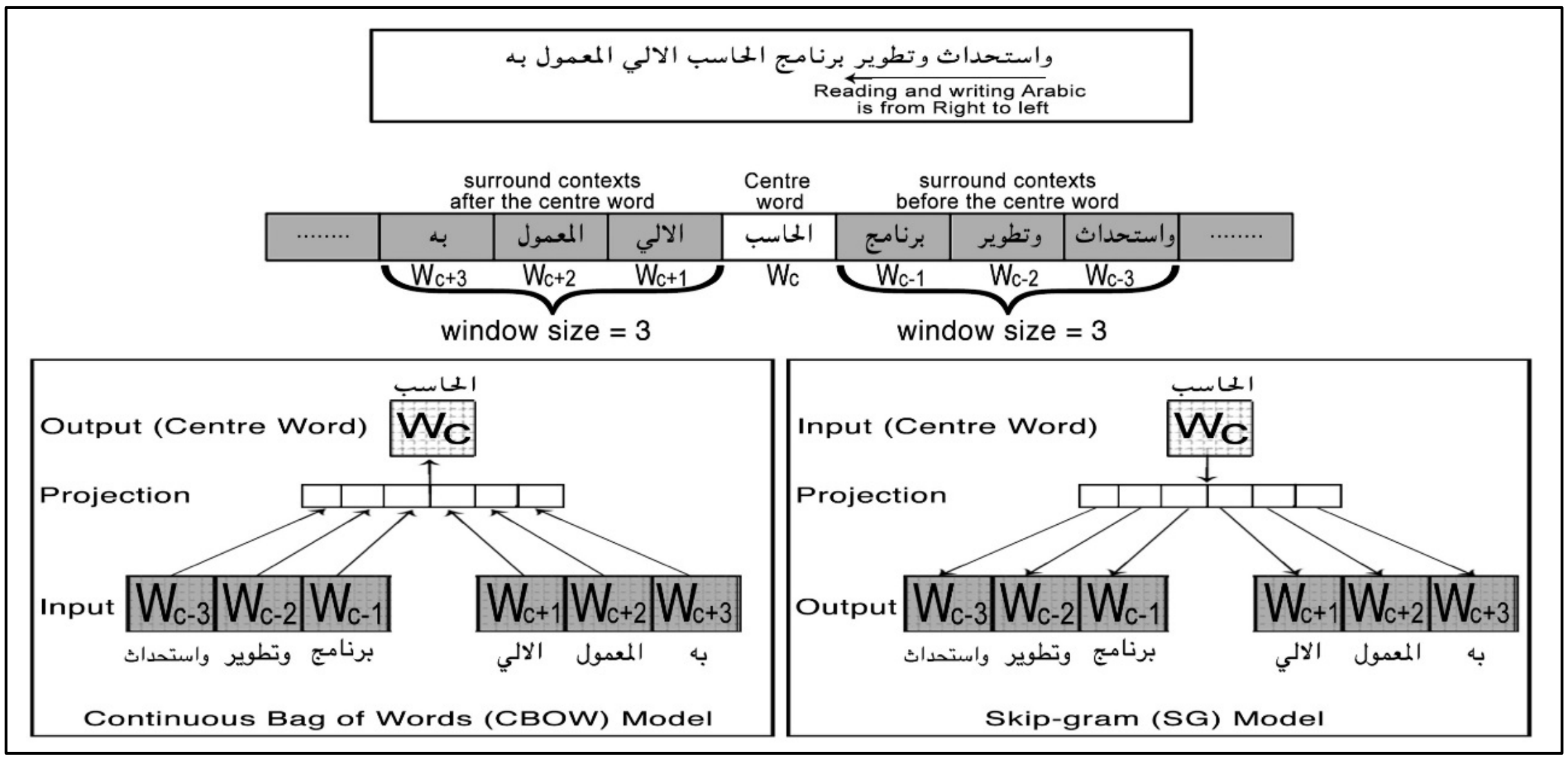

## III. ARABIC WORD EMBEDDING

Words aggregation, or mapping words that have the same meaning, is a critical issue in NLP. However, by introducing word embedding, it is no longer the semantic or syntactic word distribution that is still a crucial challenge. There are two methods that are widely used for word semantic distribution, which are Word2Vec proposed by [15], and GloVe introduced by [16]. These methods can take unstructured text and implement some mathematical equations, by representing each word in the text by a vector. All vectors that are close to each other should represent similar words.

### A. Pre-processing a Large Arabic Corpus

There are many resources from which to collect Arabic text on the web, such as Arabic Wikipedia. However, there also are Arabic corpuses that are already collected and available online. One prominent example is the Abu El-Khair Corpus that was collected by the authors of [5]. The corpus was collected from ten newspapers and it contains over three million unique words. The newspapers are from eight different Arab countries, giving the benefit to the corpus of covering of as many words in different Arabic dialects as possible.

The corpus is available in four different formats in [17] (the XML_UTF-8 format was used in this experiment). The XML tags and unwanted data, such as IDs, dates and URLs were removed, keeping the bodies of the articles and headlines only. The files were combined into a single file, and some text filtering were then applied in order to keep only the words. For example:

1. Removing any none Arabic words.
2. Removing any digit, such as 1234 or Hindi digit used in Arabic, such as ١٢٣٤.
3. Removing any special characters either in English, such as ?,;''!{}, or in Arabic, such as "؛،؟ or other common special characters, such as @#$%.
4. Normalising some letters such as ( آ ، إ ، أ ) to ا and ة to ه; etc.

The corpus size after filtering became 16.55 GB. The total number of words decreased from 1,525,722,252 to 1,520,968,919. The difference is 4,753,333 (0.31%).

### B. Building a Word2Vec Model

The Word2Vec model, proposed in [18], has two architectures: the Continuous Bag of Words (CBOW) and Skip-gram (SG). The CBOW model is to predict the centre words or the target word from the surrounding words within the windows length (see Figure I left). The SG is the opposite of CBOW and it is used to predict the surrounding words from the centre word (see Figure I right). A Gensim tool, introduced by [19], was used to implement the Word2Vec technique, and the filtered corpus was used to build an Arabic Word2Vec model.

Both architectures, i.e., CBOW and SG, were applied with different windows sizes (10, 50, 100, 200, and 300). The input to the tool is a list of sentences and each sentence has a list of words. There is a window sliding over the text to calculate the vector of each word in the corpus and distribute them in the space. The vector of each word is updated if the word appears more than once in the corpus.

### C. Choosing A Word2Vec Model

Table II in Appendix A shows twenty different results from using ten different Word2Vec models. As the models were built to be implemented as a part of the sentiment analysis, the first way to test the models is by checking the words similar to good ("جيد") and bad ("سيئ"), as these two words are the most commonly used to express positive and negative sentiment. The two approaches of Word2Vec models (CBOW and SG) were used with these two words. The Arabic letter "ء" is called Hamzah, and it is a challenging letter to be spelled correctly, if it occurs on top of or after Arabic vowel letters " ي ، و ، ا ". This problem can appear with the word for bad "سيئ" and many people might write it with incorrect spelling, such as "سيء" ، "سيىء" ، "سيئ" .

The two models (SG and CBOW) with 10 dimensions are not appropriate because the words identified as similar do not actually have the same meaning as the words for good and bad. Also, the two models SG and CBOW with 300 dimensions are not a good choice, because the CBOW has the words of opposite meaning within the top ten words.

TABLE I: THE MEAN AND THE STANDARD DEVIATION OF ALL THE CLASSIFIERS WITH DIFFERENT TEXT FEATURE SELECTIONS ON BOTH DATASETS (MAIN DATASET AND PURE DATASET)

| | Main Dataset | | | | | Sub-Dataset | | | | |
|---|---|---|---|---|---|---|---|---|---|---|
| | TF | TF-IDF | POS | Lex | Auto-Lex | TF | TF-IDF | POS | Lex | Auto-Lex |
| MNB | 0.88 (+/- 0.24) | 0.89 (+/- 0.23) | 0.89 (+/- 0.14) | 0.89 (+/- 0.17) | 0.89 (+/- 0.10) | 0.91 (+/- 0.16) | 0.92 (+/- 0.22) | 0.92 (+/- 0.17) | 0.92 (+/- 0.17) | 0.92 (+/- 0.21) |
| BNB | 0.89 (+/- 0.25) | 0.89 (+/- 0.16) | 0.89 (+/- 0.13) | 0.89 (+/- 0.17) | 0.89 (+/- 0.16) | 0.92 (+/- 0.22) | 0.93 (+/- 0.19) | 0.92 (+/- 0.13) | 0.92 (+/- 0.16) | 0.92 (+/- 0.22) |
| NSVC | 0.85 (+/- 0.27) | 0.88 (+/- 0.38) | 0.88 (+/- 0.22) | 0.88 (+/- 0.21) | 0.89 (+/- 0.24) | 0.87 (+/- 0.21) | 0.89 (+/- 0.24) | 0.89 (+/- 0.30) | 0.89 (+/- 0.12) | 0.89 (+/- 0.20) |
| LSVC | 0.90 (+/- 0.30) | 0.90 (+/- 0.26) | **0.91** (+/- 0.12) | **0.91** (+/- 0.18) | **0.91** (+/- 0.17) | **0.94** (+/- 0.12) | **0.94** (+/- 0.14) | **0.94** (+/- 0.17) | **0.94** (+/- 0.11) | **0.94** (+/- 0.22) |
| LR | 0.88 (+/- 0.24) | 0.88 (+/- 0.26) | 0.88 (+/- 0.20) | 0.88 (+/- 0.10) | 0.88 (+/- 0.15) | 0.91 (+/- 0.17) | 0.91 (+/- 0.16) | 0.91 (+/- 0.18) | 0.91 (+/- 0.16) | 0.91 (+/- 0.23) |
| SGDC | 0.89 (+/- 0.29) | 0.90 (+/- 0.24) | 0.90 (+/- 0.17) | 0.90 (+/- 0.20) | 0.90 (+/- 0.15) | 0.93 (+/- 0.16) | **0.94** (+/- 0.15) | **0.94** (+/- 0.18) | **0.94** (+/- 0.17) | **0.94** (+/- 0.19) |
| RDG | **0.91** (+/- 0.14) | **0.91** (+/- 0.13) | **0.91** (+/- 0.13) | 0.90 (+/- 0.22) | **0.91** (+/- 0.24) | **0.94** (+/- 0.14) | **0.94** (+/- 0.12) | **0.94** (+/- 0.20) | **0.94** (+/- 0.16) | **0.94** (+/- 0.19) |

The SG is suitable with the word good ,"جيد" but the similar words to bad "سيئ" are only the word bad with different Arabic spellings. In the 100 and 200 dimensions of the SG model, the opposite word bad "سيء" occurred within the list of similar word to good "جيد". The CBOW model with 50 dimensions is not an appropriate option, because of the word for "natural" or "authentic" "طبيعي" occurring as a similar word to both the words good "جيد" and bad "سيئ". The word for "natural" or "authentic" "طبيعي" cannot be classified as either positive or negative, so any models that has this word in the similar words list is not considered. Only two models have this, which are the SG with 50 dimensions and the CBOW with 100 dimensions.

As a result of this analysis, the most appropriate model to be used in this study is the CBOW with 200 dimensions.

## IV. SENTIMENT ANALYSIS

### A. Dataset

In this experiment, the Main dataset is our previously proposed dataset of Arabic tweets about health services described in [1]. The dataset was collected from Twitter and contains 628 positive tweets, and 1398 negative tweets, to give a total of 2026. As the dataset was labeled by only three human annotators, it can be hard to exactly confirm the positive or negative sentiment of each tweet, since sometimes the annotators disagreed. All the details on this can be found in [1]. We have extracted a subset from the main dataset, which we name the Sub-dataset. This contains all the tweets which all three annotators agreed as being either positive or negative. The number of positive tweets in the Sub-dataset is 502 and the number of negative tweets is 1230. So the size of the Sub-dataset is 1732 tweets (85% of the main dataset). Both datasets are freely available to download from a Bitbucket repository: [https://bitbucket.org/a_alayba/arabic-health-services-ahs-dataset/src]

### B. Sentiment classification

Previous experiments on the main dataset were described in [1], where the accuracy results were between 0.85 and 0.90 using Naïve Bayes, Support Victor Machine, Logistic Regression and Basic Deep and Convolutional Neural Networks. In this experiment, different techniques will be used, focusing on employing different feature selection methods in order to improve the accuracy. The algorithms that have been used are:

1) Multinomial Naive Bayes (MNB).
2) Bernoulli Naive Bayes (BNB).
3) Nu-Support Vector Classification(NSVC).
4) Linear Support Vector Classification(LSVC).
5) Logistic Regression (LR).
6) Stochastic Gradient Descent (SGD).
7) Ridge Classifier (RDG).

In addition, a Convolutional Neural Network has also been used.

- *Different Machine Learning Algorithms*

In this experiment, several machine learning classifiers and four different text feature selections were applied to both datasets. The text features are obtained using: Term Frequency (TF) [20], Term Frequency Inverse Document Frequency (TFIDF) [20], Part of Speech tagging (POS) [20], a manual built lexicon (Lex) and an Automatic Lexicon (Auto-Lex). The TF is the frequency of each word in the corpus. The TFIDF is obtained by weighting each word in the corpus, by combing the frequency of the word and the inverse document frequency. The POS for Arabic was generated using the Stanford CoreNLP toolkit [21]. Consider the sentence "الخدمات الصحيه الحاليه متدهوره و بحاجه الى تغيير الادارات", whose English translation is "The current health services are deteriorating and they need to change the departments' administrations". After applying the Part of Speech tagging, the sentence will be:

((الخدمات, DTNNS) (الصحيه, DTJJ) (ه, PRP) (الحالي, DTJJ) (ه, PRP) (متدهور, JJ) (ه, PRP) (و, CC) (ب, IN) (حاجه, NN) (الى, IN) (تغيير, NN) (الادارات, DTNNS)).

In the experiment, the focus was only on Verbs (VBD) and Adjectives (JJ). The Lexicon is built manually by collecting the most common positive and negative words in the corpus. The Auto-Lex is automatically collected by using the Word2Vec model from the Abu El-Khair Arabic Corpus. The initial words are (good "جيد" and bad "سيئ"), which are the most common words in opinion/sentiment analysis. Based on these words, the function "`most_similar`" from the Genism tool [19] was used to retrieve the nearest ten words to each of them. After that, we expand the lexicon by generating the five most similar words of each word from the first result. The reason for choosing only five words is to avoid adding any opposite words. Finally, we removed any duplicated words.

FIGURE II: THE ACCURACY ON THE TRAINING AND TESTING SET FOR BOTH DATASETS, THE MAIN AND THE SUB-DATASETS, USING THREE WAYS OF LEXICONS (SEMEVAL-2016 ARABIC TWITTER LEXICON, ARABIC HEALTH TWITTER LEXICON AND A COMBINATION OF BOTH)

| Lexicons | The Main Dataset | The Sub-Dataset |
|---|---|---|
| **SemEval-2016 Arabic Twitter Lexicon** | 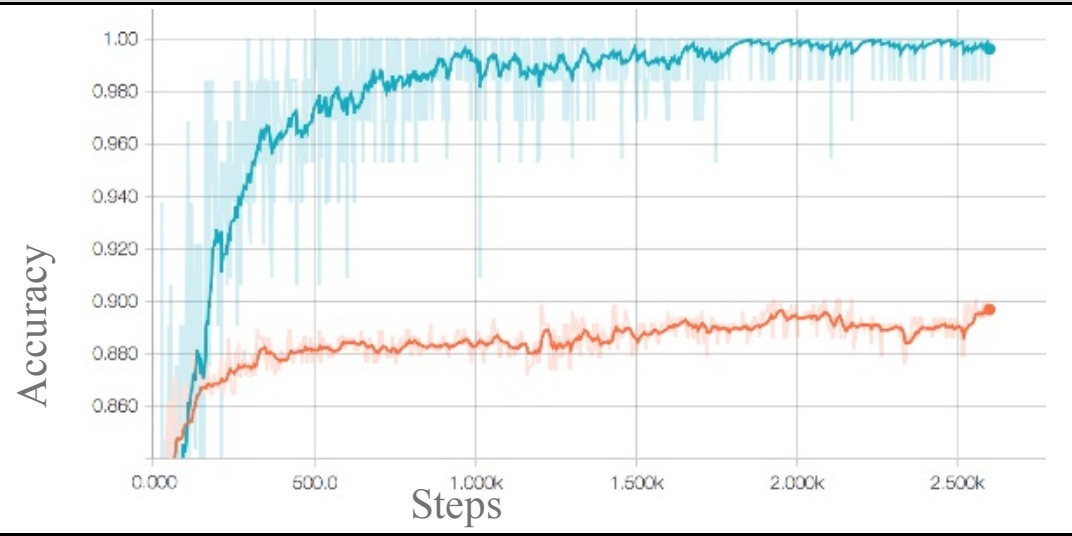  | 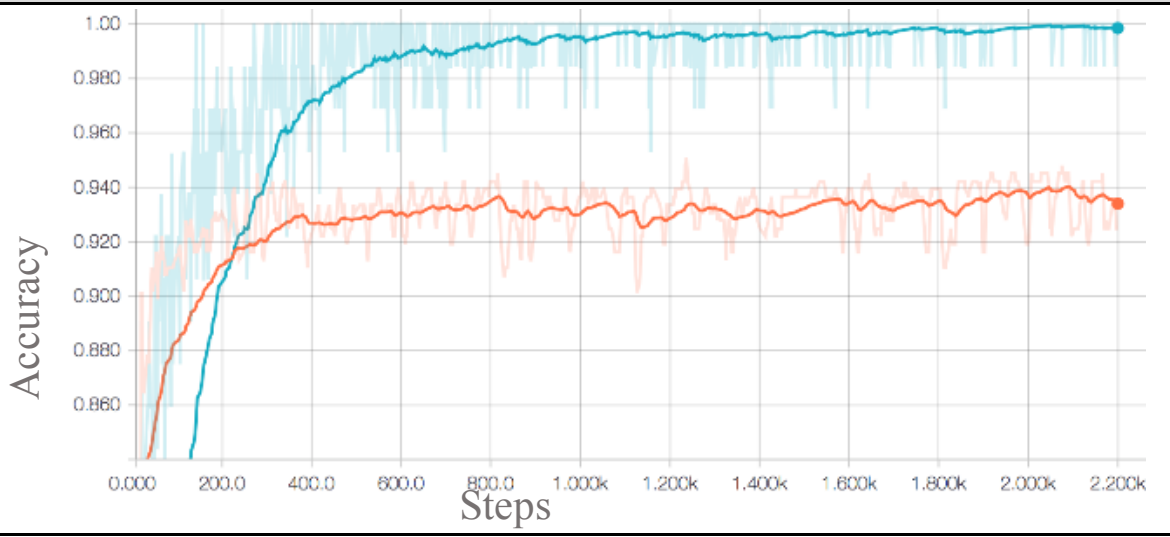  |
| **Arabic Health Twitter Lexicon** | 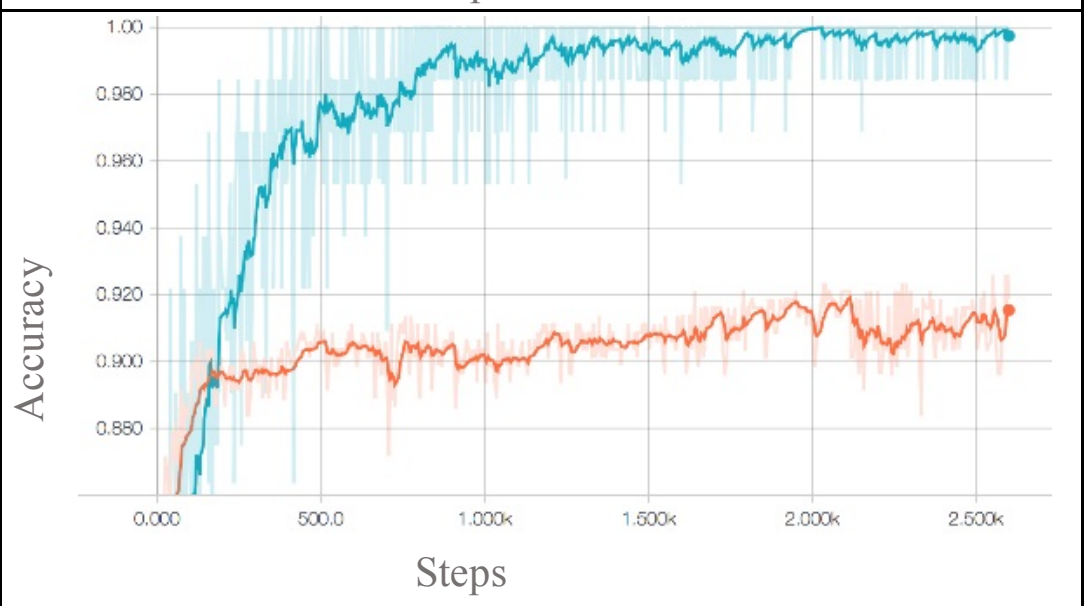  | 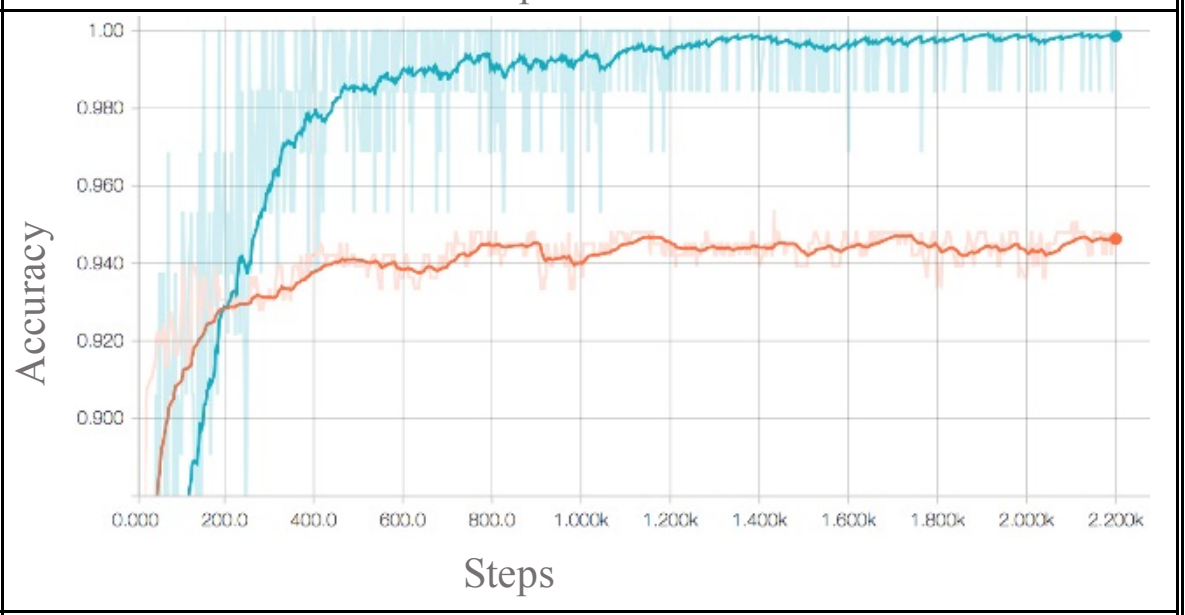  |
| **Both** | 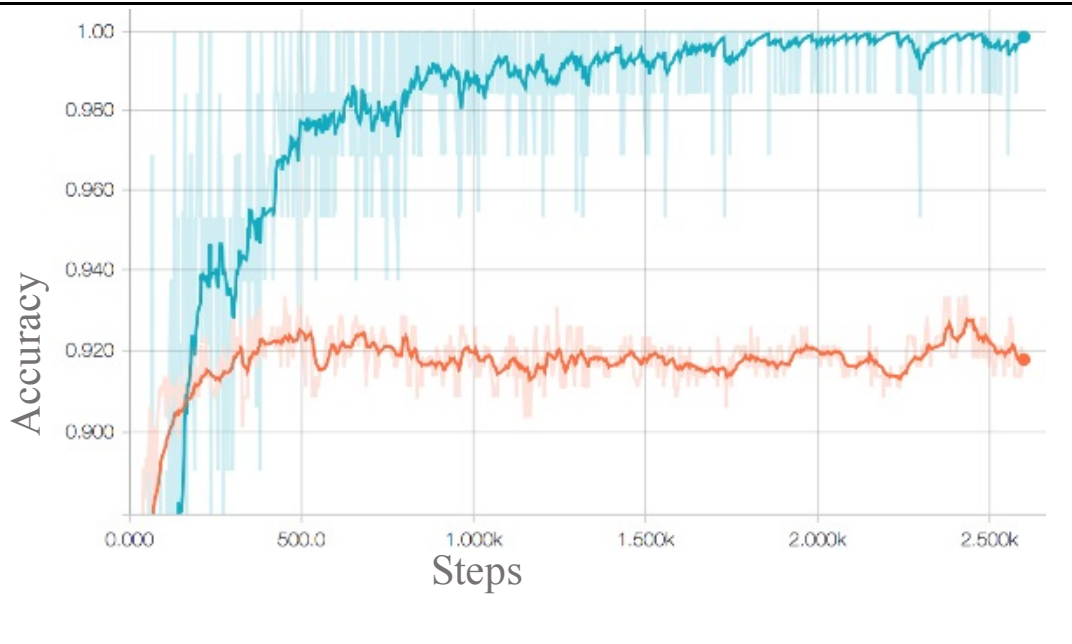  | 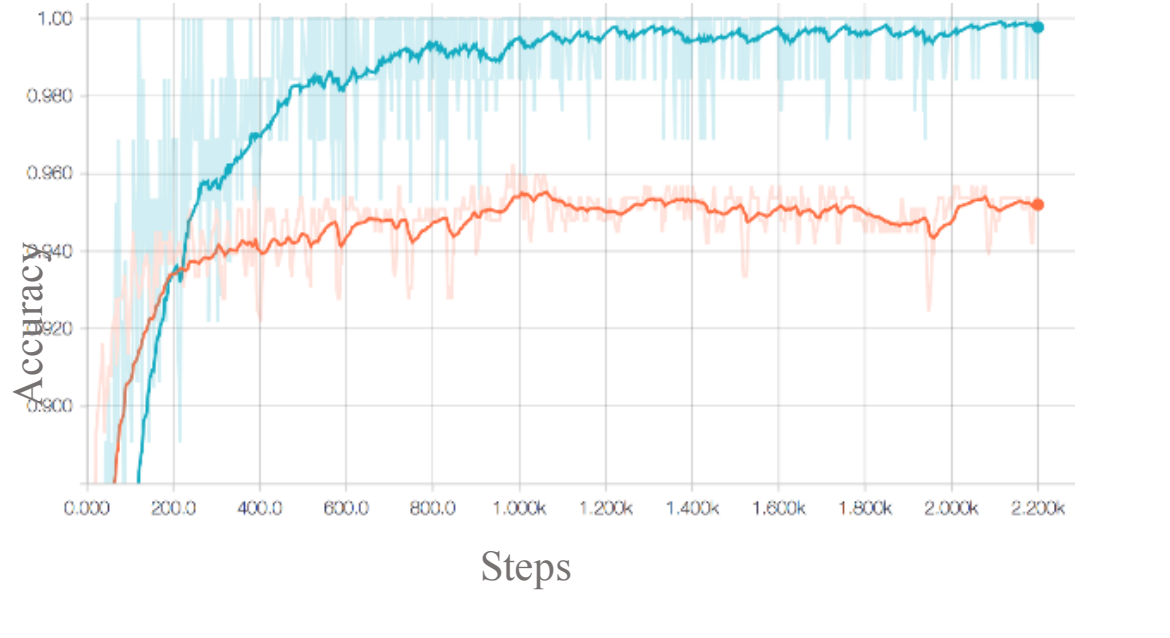  |
| | ● The accuracy on the training set | ● The accuracy on the testing set |

Both lexicons have two values: +1 for positive words and -1 for negative words. The Pipeline in the Scikit Learn tool [22] was used in order to implement the sentiment classification using different text feature selections. In the experiment, cross validation was used and we calculated the mean in a ten-fold cross validation to get reliable results. Table I shows all the results of using different algorithms and different features.

The results are the accuracies and standard deviations obtained using both the Main dataset and the Sub-dataset. Linear Support Vector Classification and Ridge Classifier shows the best results on both datasets.

- *Convolutional Deep Neural Networks*

Convolutional Neural Networks (CNNs) are a powerful method and show very good results in natural language processing. There has been much NLP research that used CNNs, such as [23], [24] and [13]. In particular, [25] presented an integrated CNN and Lexicon models, with one of them called Naïve Concatenation. Some modifications were applied to this model in order to implement the sentiment analysis on our Arabic Health Services dataset. Two Arabic lexicons were used in this experiment: the SemEval-2016 Arabic Twitter Lexicon [26] and one which was built manually based on our Arabic Health dataset (Arabic Health Twitter Lexicon). The SemEval-2016 Arabic Twitter Lexicon contains 1366 words and the values scales are between +1.0 and -1.0. The Arabic Health Twitter Lexicon contains 716 words and it uses only two values: either -1 or +1. The word embedding model, which is the CBOW with 200 dimensions, was used in order to expand the lexicon and improve the classification. The filter sizes are (3, 4 and 5), the filter size for the Word2Vec model is 64 and the filter size for the lexicon is 9. The numbers of evaluation epochs was 100 and the dataset is divided into 80% for training and 20% for testing.

This model was applied on both datasets, i.e., the Main dataset and the Sub-dataset. The accuracy achieved was 0.92 for the Main dataset using both lexicons and the accuracy rises to 0.95 for the Sub-dataset using both lexicons as well. Figure II illustrates the accuracy of the sentiment analysis using this approach, which has increased from 0.88 to 0.92 on the Main dataset. Also, the accuracy of the Sub-dataset is between 0.93 and 0.95.

## V. Conclusion

This paper exploits the benefit of word embedding by using Word2Vec in order to gain similar words. In addition, we have used a 1.5 billon words corpus (Abu El-Khair Corpus) in order to involve as many words as possible and different Arabic dialects. The paper explains the task of pre-processing the large Arabic corpus, constructing the Word2Vec models and selecting the best model. The best Word2Vec model was used to build an Automatic Arabic Lexicon that used with different Machine Learning methods. Also, it has been used apart from of the lexicon in Convolutional Neural Networks in order to expand the vocabularies. These approaches have increased the sentiment classification for our Arabic Health Services dataset (AHS) from 0.85 to 0.92 for the Main dataset, and from 0.87 to 0.95 for the Sub-dataset. Finally, this paper presents an improved accuracy, reaching 0.92, compared to our previous results in [1] that were 0.90 on the Main-Dataset.

We plan future studies to deal with negation words in Arabic, as the negative or the opposite word meaning might be avoided by viewing it as a compound word of two words.

## Appendix A:

TABLE II. THE RESULTS OF THE MOST SIMILAR WORDS TO GOOD AND BAD IN ARABIC USING THE TWO TECHNIQUES CBOW AND SK WITH DIFFERENT DIMENSIONALITY (10, 50, 100, 200, AND 300)

| Models | Dimensionality | Good "جيد" | Bad "سيئ" |
|---|---|---|---|
| Continuous Bag of Word (CBOW) | 10 | CBOW_model_10.most_similar("جيد", topn=10)<br>1: (0.9796178340911865 ,'مردودهم') Profit + NA<br>2: (0.9740296602249146 ,'افتقاده') Missing<br>3: (0.9646542072296143 ,'التمرير') Passing<br>4: (0.9547749161720276 ,'تركيزه') Concentration<br>5: (0.9546589255332947 ,'الريتم') Rhythm<br>6: (0.9528429508209229 ,'ادائه') Performance<br>7: (0.9487570524215698 ,'مهيا') Prepared<br>8: (0.9436195492744446 ,'احترافي') Professional<br>9: (0.9425039887428284 ,'الاسترجاع') Refund / Returned<br>10: (0.9403938055038452 ,'ادائهما') Performance | CBOW_model_10.most_similar("سيئ", topn=10)<br>1: (0.9720628261566162 ,'سيء') Bad<br>2: (0.9698752164840698 ,'سيىء') Bad<br>3: (0.956177830696106 ,'سىء') Bad<br>4: (0.9537068605422974 ,'مرتبك') Confused<br>5: (0.9514274001121521 ,'ريالنا') {NA}<br>6: (0.9502087235450745 ,'سئ') Bad<br>7: (0.9434422850608826 ,'ومتكررا') Repeated<br>8: (0.9427366852760315 ,'مشلولا') Paralysed<br>9: (0.9408395290374756 ,'عكسي') Opposite / Inverse<br>10: (0.9399359822273254 ,'بخسا') Cheap / {NA} |
| | 50 | CBOW_model_50.most_similar("جيد", topn=10)<br>1: (0.8463985323905945 ,'مثالي') Ideal<br>2: (0.8301881551742554 ,'ممتاز') Excellent / Perfect<br>3: (0.828335165977478 ,'مريح') Comfortable<br>4: (0.8220809698104858 ,'طبيعي') Natural<br>5: (0.8217611312866211 ,'متواضع') Modest<br>6: (0.8159613013267517 ,'رائع') Wonderful / Marvelous<br>7: (0.8152831196784973 ,'طبيعى') Natural<br>8: (0.8143441081047058 ,'كافي') Enough<br>9: (0.8141897916793823 ,'مميز') Distinctive / Special<br>10: (0.8072431087493896 ,'احترافي') Professional | CBOW_model_50.most_similar("سيئ", topn=10)<br>1: (0.9738459587097168 ,'سيىء') Bad<br>2: (0.966951310634613 ,'سيء') Bad<br>3: (0.9502550363540649 ,'سيىء') Bad<br>4: (0.9420933723449707 ,'سىء') Bad<br>5: (0.9238800406455994 ,'سئ') Bad<br>6: (0.8482351899147034 ,'مزعج') Annoying<br>7: (0.8412387371063232 ,'خاطيء') Erroneous / Wrong<br>8: (0.840319037437439 ,'طبيعي') Natural / Authentic<br>9: (0.8346652984619141 ,'مقلق') Worrying<br>10: (0.8317472338676453 ,'خاطىء') Erroneous / Wrong |
| | 100 | CBOW_model_100.most_similar("جيد", topn=10)<br>1: (0.8275733590126038 ,'مثالي') Ideal<br>2: (0.8227100372314453 ,'ممتاز') Excellent / Perfect<br>3: (0.803859293460846 ,'احترافي') Professional<br>4: (0.7994316816329956 ,'متواضع') Modest<br>5: (0.7944766283035278 ,'رائع') Wonderful / Marvelous<br>6: (0.792463481426239 ,'مميز') Distinctive / Special<br>7: (0.786441445350647 ,'طبيعي') Natural<br>8: (0.785868644714355 ,'مريح') Comfortable<br>9: (0.7804467678070068 ,'طبيعى') Natural<br>10: (0.7740051746368408 ,'متميز') Distinctive / Special | CBOW_model_100.most_similar("سيئ", topn=10)<br>1: (0.9771776795387268 ,'سيىء') Bad<br>2: (0.9742478132247925 ,'سيء') Bad<br>3: (0.9475470781326294 ,'سىء') Bad<br>4: (0.9123654961585999 ,'سيىء') Bad<br>5: (0.9029320478439331 ,'سئ') Bad<br>6: (0.7654193639755249 ,'مقلق') Worrying<br>7: (0.7645547986030579 ,'مزعج') Annoying<br>8: (0.7608690857887268 ,'كارثي') Disastrous<br>9: (0.7600266933441162 ,'خاطىء') Erroneous / Wrong<br>10: (0.7579468488693237 ,'مقزز') Disgusting |
| | 200 | CBOW_model_200.most_similar("جيد", topn=10)<br>1: (0.7963066101074219 ,'ممتاز') Excellent / Perfect<br>2: (0.7615153193473816 ,'رائع') Wonderful / Marvelous<br>3: (0.7590458393096924 ,'مميز') Distinctive / Special<br>4: (0.7560107707977295 ,'مثالي') Ideal<br>5: (0.7487468719482422 ,'متميز') Distinctive / Special<br>6: (0.7445393800735474 ,'متواضع') Comfortable<br>7: (0.7248423099517822 ,'مريح') Modest<br>8: (0.7046573162078857 ,'قوي') Strong<br>9: (0.6964720487594604 ,'مثالى') Ideal<br>10: (0.6958665251731873 ,'احترافي') Profissional | CBOW_model_200.most_similar("سيئ", topn=10)<br>1: (0.9636447429656982 ,'سيء') Bad<br>2: (0.9589278101921082 ,'سيىء') Bad<br>3: (0.928744912147522 ,'سىء') Bad<br>4: (0.8867411613464355 ,'سيىء') Bad<br>5: (0.8663808107376099 ,'سئ') Bad<br>6: (0.7519478797912598 ,'كارثي') Disastrous<br>7: (0.7462688088417053 ,'مزر') Miserable<br>8: (0.7351321578025818 ,'ماسوي') Tragic<br>9: (0.7105754613876343 ,'كارثى') Disastrous<br>10: (0.7087178230285645 ,'مزعج') Annoying |
| | 300 | CBOW_model_300.most_similar("جيد", topn=10)<br>1: (0.7638713717460632 ,'ممتاز') Excellent / Perfect<br>2: (0.7417177557945251 ,'رائع') Wonderful / Marvelous<br>3: (0.7398247718811035 ,'مميز') Distinctive / Special<br>4: (0.7240325212478638 ,'متميز') Distinctive / Special<br>5: (0.7061358690261841 ,'مثالي') Ideal<br>6: (0.7055904865264893 ,'متواضع') Modest<br>7: (0.6876623034477234 ,'مريح') Comfortable<br>8: (0.6684661507606506 ,'سيىء') Bad<br>9: (0.6675729155540466 ,'سيء') Bad<br>10: (0.6646686792373657 ,'مثالى') Ideal | CBOW_model_300.most_similar("سيئ", topn=10)<br>1: (0.9508958458900452 ,'سيء') Bad<br>2: (0.9418485760688782 ,'سيىء') Bad<br>3: (0.893813967704773 ,'سىء') Bad<br>4: (0.8653604984283447 ,'سيىء') Bad<br>5: (0.8139645457267761 ,'سئ') Bad<br>6: (0.7065134048461914 ,'كارثي') Disastrous<br>7: (0.6920626759529114 ,'مزر') Miserable<br>8: (0.6803059577941895 ,'كارثى') Disastrous<br>9: (0.671885073184967 ,'ماسوي') Tragic<br>10: (0.6675729155540466 ,'جيد') Good |
| Skip-gram (SG) | 10 | SG_model_10.most_similar("جيد", topn=10)∂)<br>1: (0.9871622920036316 ,'لمناسب') Appropriate / Suitable<br>2: (0.9809944033622742 ,'سيتناسب') Will fit<br>3: (0.9765548706054688 ,'تسنى') Will be able<br>4: (0.9678409695625305 ,'الاسبقيه') Priority / Precedence<br>5: (0.9674731492996216 ,'لتحفيزه') Motivate / Encourage<br>6: (0.967185378074646 ,'استفادته') Profit / Benefit<br>7: (0.9635581374168396 ,'لمعاملاته') Dealing / Treatment<br>8: (0.9617605209350586 ,'للمستوي') To the level<br>9: (0.9608638882637024 ,'مهيا') Prepared / Ready<br>10: (0.9596813321113586 ,'تمتعه') Joy | SG_model_10.most_similar("سيئ", topn=10)<br>1: (0.9950416088104248 ,'سيىء') Bad<br>2: (0.9876136183738708 ,'سيتخلص') Will get Rid<br>3: (0.9848575592041016 ,'سيىء') Bad<br>4: (0.9776631593704224 ,'سيء') Bad<br>5: (0.9757347702980042 ,'للحراق') {NA}<br>6: (0.9734814167022705 ,'المفاجا') {NA}<br>7: (0.9734290838241577 ,'فقط') Just / Only<br>8: (0.9698436260223389 ,'تلاشيه') To fade<br>9: (0.9690333008766174 ,'وسيتركه') Left it<br>10: (0.9684392213821411 ,'وعاجزا') Unable / Disable |
| | 50 | SG_model_50.most_similar("جيد", topn=10) )<br>1: (0.8267565369606018 ,'طبيعى') Natural / Authentic<br>2: (0.8181592226028442 ,'بشكل') In a form<br>3: (0.805439829826355 ,'احترافي') Professional<br>4: (0.7810139656066895 ,'وجيد') And good<br>5: (0.7771871089935303 ,'متميز') Distinct / Special<br>6: (0.7686337828636169 ,'ايجابي') Positive<br>7: (0.7644679546356201 ,'متدنى') Low<br>8: (0.7636167407035828 ,'اساسى') Basic / Essential<br>9: (0.7575806379318237 ,'مثالى') Ideal<br>10: (0.7490912675857544 ,'واقعى') Realistic | SG_model_50.most_similar("سيئ", topn=10)<br>1: (0.9751160144805908 ,'سيىء') Bad<br>2: (0.9571614265441895 ,'سيء') Bad<br>3: (0.9245176315307617 ,'سيىء') Bad<br>4: (0.8288180232048035 ,'سىء') Bad<br>5: (0.820361852645874 ,'سئ') Bad<br>6: (0.8165686726570129 ,'وسيئ') And bad<br>7: (0.8098408579826355 ,'وسيء') And bad<br>8: (0.8015778064727783 ,'ومزر') And Miserable<br>9: (0.7994084358215332 ,'مقلق') Worrying<br>10: (0.7965108156204224 ,'مزعج') Annoying |
| | 100 | SG_model_100.most_similar("جيد", topn=10))<br>1: (0.781434178352356 ,'وجيد') And good<br>2: (0.7575291395187378 ,'متميز') Distinct / Special<br>3: (0.7511346936225891 ,'بشكل') In a form<br>4: (0.7370492219924927 ,'طبيعى') Natural / Authentic<br>5: (0.7258888483047485 ,'جيده') Good<br>6: (0.7172396779060364 ,'مثالى') Ideal<br>7: (0.7065442800521851 ,'ايجابي') Positive<br>8: (0.7054450511932373 ,'احترافي') Professional<br>9: (0.7051820755004883 ,'سىء') Bad<br>10: (0.7044427990913391 ,'واضح') Clear | SG_model_100.most_similar("سيئ", topn=10)<br>1: (0.9562578201293945 ,'سيىء') Bad<br>2: (0.9476160407066345 ,'سيء') Bad<br>3: (0.9023429751396179 ,'سيىء') Bad<br>4: (0.8343167304992676 ,'سىء') Bad<br>5: (0.8138883113861084 ,'وسيئ') And bad<br>6: (0.8057430982589722 ,'سئ') Bad<br>7: (0.7410703897476196 ,'وسيء') And bad<br>8: (0.7346179485321045 ,'كارثي') Disastrous<br>9: (0.7293833494186401 ,'مؤسف') Regrettable<br>10: (0.7200167775154114 ,'مقلق') Worrying |
| | 200 | SG_model_200.most_similar("جيد", topn=10))<br>1: (0.7182195782661438 ,'متميز') Distinct / Special<br>2: (0.7180838584899902 ,'وجيد') And good<br>3: (0.661284327507019 ,'جيده') Good<br>4: (0.654598593711853 ,'رائع') Wonderful / Marvelous<br>5: (0.646915078163147 ,'مناسب') Appropriate / Suitable<br>6: (0.6466327905654907 ,'مميز') Distinct / Special<br>7: (0.645879864692688 ,'واضح') Clear<br>8: (0.6426805257797241 ,'طبيعى') Natural / Authentic<br>9: (0.6356602311134338 ,'سىء') Bad<br>10: (0.6325259804725647 ,'مثالى') Ideal | SG_model_200.most_similar("سيئ", topn=10)<br>1: (0.9353238344192505 ,'سيء') Bad<br>2: (0.9160232543945312 ,'سيىء') Bad<br>3: (0.7935423851013184 ,'سيىء') Bad<br>4: (0.7717869281768799 ,'سىء') Bad<br>5: (0.7513115406036377 ,'سئ') Bad<br>6: (0.7098520994186401 ,'السيء') The bad<br>7: (0.7028955221176147 ,'وسيئ') And bad<br>8: (0.69720858335495 ,'وسيء') And bad<br>9: (0.6894170045852661 ,'سيئا') Bad<br>10: (0.6735714673995972 ,'مؤسف') Regrettable |
| | 300 | SG_model_300.most_similar("جيد", topn=10))<br>1: (0.7157446146011353 ,'وجيد') And good<br>2: (0.6670716404914856 ,'متميز') Distinct / Special<br>3: (0.6348607540130615 ,'مميز') Distinct / Special<br>4: (0.6342040300369263 ,'رائع') Wonderful / Marvelous<br>5: (0.6238257884979248 ,'مناسب') Appropriate / Suitable<br>6: (0.619631290435791 ,'ممتاز') Excellent / Perfect<br>7: (0.6119194030761719 ,'جيده') Good<br>8: (0.6048153638839722 ,'واضح') Clear<br>9: (0.5936441421508789 ,'جيدا') Good<br>10: (0.5928791165351868 ,'طبيعى') Natural / Authentic | SG_model_300.most_similar("سيئ", topn=10)<br>1: (0.9151774048805237 ,'سيء') Bad<br>2: (0.8923256993293762 ,'سيىء') Bad<br>3: (0.769922137260437 ,'سيىء') Bad<br>4: (0.7417573928833008 ,'سىء') Bad<br>5: (0.695048987865448 ,'سئ') Bad<br>6: (0.683836042881012 ,'وسيئ') And bad<br>7: (0.6823018193244934 ,'سيئا') Bad<br>8: (0.6465323567390442 ,'السيء') The bad<br>9: (0.6368034482002258 ,'وسيء') And bad<br>10: (0.6225597858428955 ,'السيىء') The bad |